\documentclass[letterpaper, 10 pt, conference]{ieeeconf}  

\IEEEoverridecommandlockouts                              

\usepackage{amsmath}
\usepackage{cite}
\usepackage{amssymb}
\usepackage{xcolor}
\usepackage{xspace}
\usepackage{amsmath}
\usepackage{graphicx} 
\usepackage{booktabs}
\usepackage{multirow}
\usepackage{multicol}
\usepackage{makecell}
\usepackage{amsfonts}
\usepackage{amssymb}
\usepackage{pifont}
\usepackage{graphicx}
\usepackage{threeparttable}
\usepackage{wrapfig}
\usepackage{hyperref}
\newcommand{\hlc}[1]{#1}
\newcommand{\ours}{CARF\xspace}

\title{\LARGE \bf
CARF: \textbf{C}ontrastive \textbf{A}ttraction–\textbf{R}epulsion of \textbf{F}ailure-Guided Flow Matching
}

\author{Shuqi Zhao, Bang Du, Cheng-en Wu, Yichen Xie, Yixiao Wang, Masayoshi Tomizuka}

\begin{document}

\maketitle
\thispagestyle{empty}
\pagestyle{empty}

\begin{abstract}

Robot demonstration collection often produces imperfect or failed trajectories in addition to successful demonstrations. Existing methods typically exploit failed trajectories by identifying segments that still make progress toward task completion, but largely overlook \textit{failure-critical behaviors} that directly lead to task failure. Here we argue that these two types of segments provide fundamentally asymmetric supervision: progressive segments should be imitated, whereas failure-critical segments should be explicitly avoided. Based on this observation, we propose \textbf{CARF}, a \textbf{C}ontrastive \textbf{A}ttraction-\textbf{R}epulsion of \textbf{F}ailure-guided framework for learning from imperfect robot data. 
CARF introduces a progress-based importance scorer, trained solely on successful expert demonstrations and its perturbation results, to estimate step-wise contributions toward task completion and identify informative regions in failed trajectories. These scores guide a unified flow-matching objective that attracts the policy toward progressive behaviors and repels it from failure-critical ones, while excluding ambiguous segments. This enables more comprehensive utilization of imperfect data and avoids unreliable supervision from ambiguous failure segments. Extensive experiments in simulation and the real world demonstrate consistent improvements over competing baselines across diverse failure scenarios, with ablations further validating the effectiveness of the proposed scoring and attraction-repulsion mechanisms. Our website is \url{https://zhao-sq.github.io/carf/#}.
\end{abstract}

\section{INTRODUCTION}
\label{sec:intro}

Different from images or texts easily available from the Internet, the collection of robot data has always been a serious challenge due to its extremely time-consuming procedure and intensive human labor~\cite{chi2024universal,xu2025dexumi,fang2025dexop,zhao2025dexh2r,mao2026beyond,zheng2026rewind}. 
Consequently, it is of great necessity for modern robot learning to make full use of all available robot data ~\cite{xie2026multi,chen2026craft,kim2021demodice,huang2025fail2progress}, especially given that the human demonstration collection process inevitably involves not only successful but also imperfect or failed task executions~\cite{robomimic2021}.
Recent efforts have explored different ways to leverage imperfect robot data. Offline reinforcement learning is one possible direction, where policies are improved from fixed datasets through value estimation under sparse success or failure rewards~\cite{huang2025using,intelligence2025pi,giridhar2026beyond}. However, such methods often depend on indirect long-horizon credit assignment through bootstrapped value propagation, which can make training unstable in practice. Another line of imitation learning methods selects useful suboptimal data based on feature similarity and reuses selected segments as additional demonstrations~\cite{wu2025learning,lin2024flowretrieval,du2023behavior}. 


However, existing methods primarily focus on identifying \textit{Progressive Segments} from imperfect data, i.e., segments that still make meaningful progress toward task completion, while largely overlooking the information contained in the remaining portions. In fact, these remaining segments can be further divided into two categories: \textit{Indeterminate Segments} and \textit{Failure-Critical Segments}. While \textit{Indeterminate Segments} provide ambiguous supervision due to their mixed execution quality, \textit{Failure-Critical Segments} can still be effectively exploited, as they correspond to decisive erroneous behaviors that directly prevent task success. Fundamentally, \textit{Progressive Segments} and \textit{Failure-Critical Segments} provide asymmetric learning signals: a policy should imitate the former while avoiding the latter, while focusing solely on \textit{Progressive Segments} may leave valuable information in the remaining robot data underutilized.
\begin{figure}[t]
  \centering
   \includegraphics[width=\columnwidth]{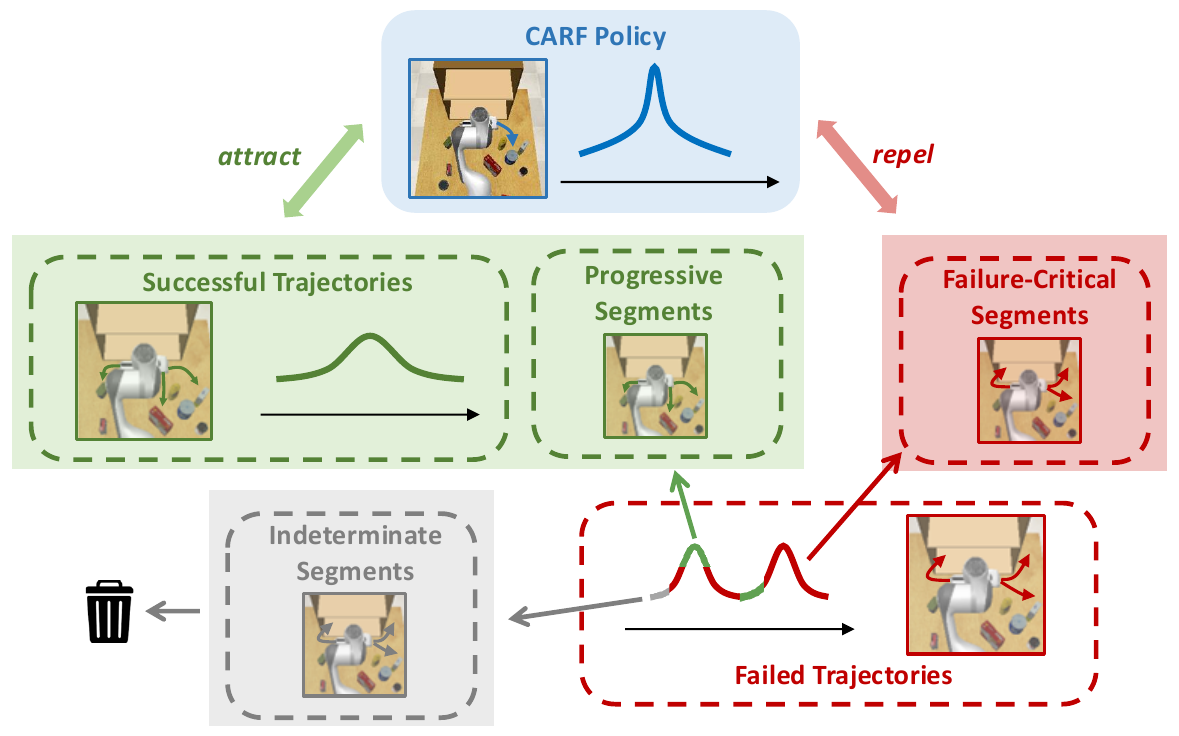}
   \caption{CARF leverages a crucial yet largely underexplored source of data: \textit{Failure-Critical Segments}. It further formulates an attraction-repulsion flow-matching policy that pulls the policy toward the correct action distribution while pushing it away from failure distributions.}
   \label{fig:teaser}
\end{figure}

To this end, we propose CARF, a framework that explicitly models these asymmetric learning signals through attraction and repulsion. 
To achieve this, we introduce a progress-based importance scorer learned solely from successful expert trajectories to quantify the contribution of each step's state-action pairs to the final task completion, providing a step-wise usefulness estimate in failed task executions. Both successful and failure data are then integrated into a unified training process, where the frozen trained scorer decides the asymmetric attraction–repulsion behavior of the flow dynamics for each step-wise action. This explicitly steers the flow matching policy towards success and away from failure. Compared to previous methods, our approach enables more comprehensive utilization of available robot data by extracting complementary learning signals from both progressive and failure-critical behaviors. Overall, our contributions are listed as follows:
\begin{itemize}
    \item We propose \hlc{\textbf{CARF}, a novel \textbf{C}ontrastive \textbf{A}ttraction-\textbf{R}epulsion of \textbf{F}ailure-guided framework} that enables policy to systematically extract informative signals from both \textit{Progressive Segments} and \textit{Failure-Critical Segments} action clips, enabling more comprehensive data utilization.
    \item We introduce a progress-based importance scoring mechanism that explicitly quantifies step-wise contribution to task success and integrate it into an attraction–repulsion strategy, leading to more stable and targeted policy learning.
    \item We conduct extensive experiments with thorough ablations and visualizations to validate our methods in both simulation and real world. Our method outperforms all proposed baselines in multiple types of failure scenarios.
\end{itemize}

\section{RELATED WORK}
\label{sec:rw}
\textbf{Imitation Learning}
Imitation learning provides a simple and effective paradigm for visuomotor policy learning by directly fitting policies to expert demonstrations~\cite{zheng2024data,xia2025cage,zhao2025dexctrl, zheng2026egoscale, xu2025seeing}. Early behavior cloning methods commonly predict single-step actions, while recent sequence-modeling approaches such as ACT~\cite{zhao2023learning} use action chunking and transformer to model temporally coherent action sequences. Generative policies further improves the expressiveness of imitation learning: Diffusion Policy~\cite{chi2025diffusion} represents action trajectories through an iterative denoising process, enabling multimodal action prediction under image observations, while flow-matching-based policies learn continuous vector fields that transport noise toward expert action distributions with more direct training objectives~\cite{lipman2022flow,stoica2025contrastive}. Despite these advances, most imitation learning methods still rely primarily on successful demonstrations and treat failed rollouts as unusable or harmful. Instead, our method extends flow-based imitation learning to imperfect datasets by explicitly modeling both attractive and repulsive signals, allowing visuomotor policies to learn from broader robot experience.

\textbf{Learning from failure}
A natural way to improve data efficiency in imitation learning is to reuse imperfect or failed trajectories rather than discarding them entirely~\cite{xu2022discriminator,huang2025using, dass2025datamil, wei2026ambient, zheng2026failing, hao2026far,yan2026progressvla}. Existing methods commonly estimate the usefulness of such data through classifier guidance~\cite{dhariwal2021diffusion}, trajectory-quality prediction, or similarity-based scoring between failed segments and successful demonstrations~\cite{wu2025learning,lin2024flowretrieval,du2023behavior}. Related to this goal, offline reinforcement learning also learns from fixed datasets without additional environment interaction~\cite{kumar2020conservative,kostrikov2021offline,hansen2023idql}, but many representative methods are primarily evaluated in state-based settings, while recent visuomotor extensions often involve human-in-the-loop correction during training~\cite{intelligence2025pi}. Similarity-based reuse can be misleading: near-success segments may have high similarity to successful demonstrations, yet still encode critical failure modes that the final policy should avoid. Our method addresses this limitation with an attraction--repulsion flow-matching objective, which attracts the policy toward successful behaviors while explicitly repels it from near-success-but-still-failed behaviors, enabling a simple imitation-learning pipeline to benefit from imperfect visuomotor data without online interaction or human intervention during training.
\begin{figure*}[t]
  \centering
   \includegraphics[width=\textwidth]{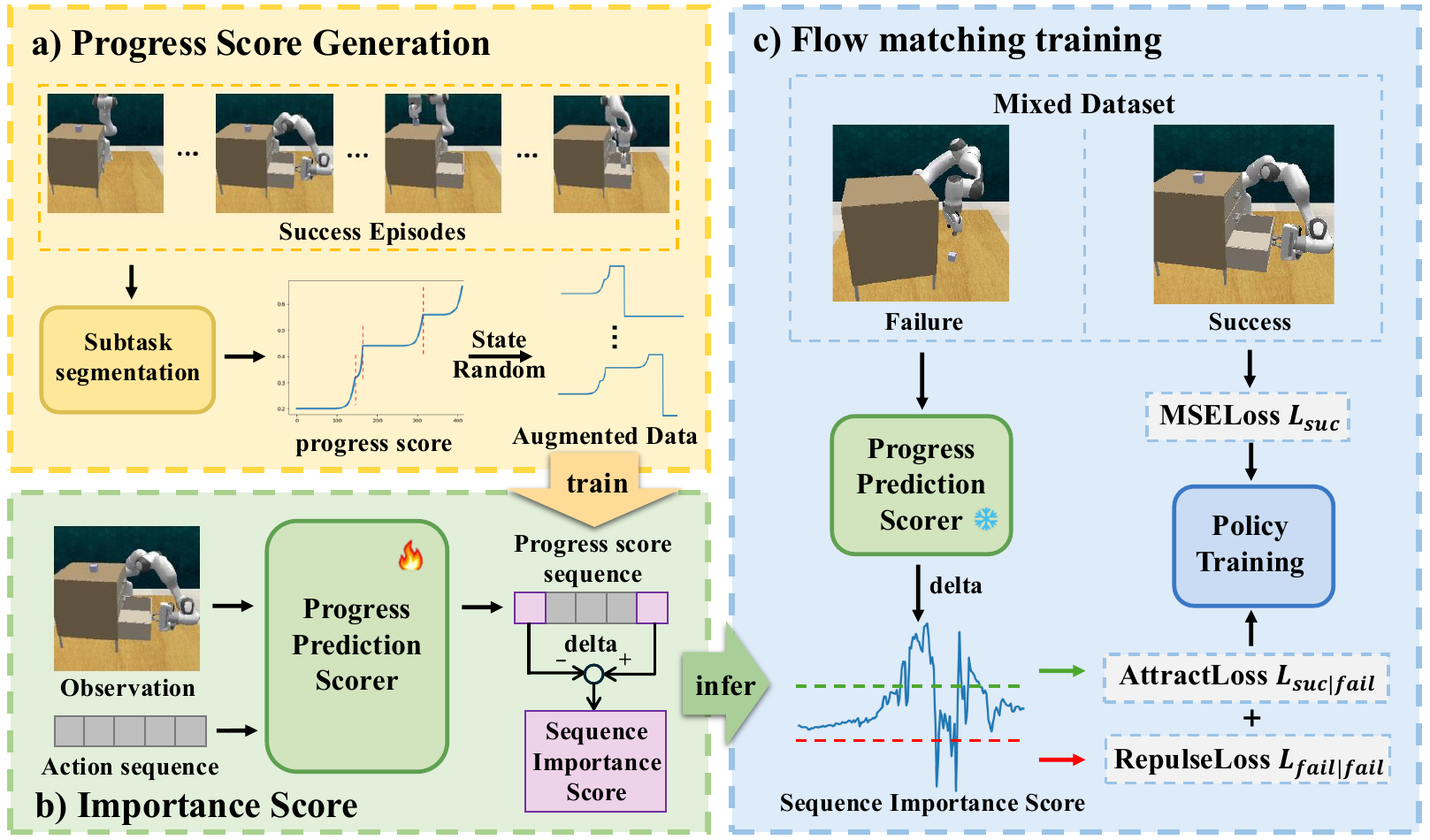}
   \vspace{-2pt}
   \caption{Overview framework of \ours. After automatically labeling success dataset based on subtask segmentation, \ours trains a progress prediction scorer that provides importance score for observation-action chunking sequence, which is used for attraction-repulsion flow matching training.
   }
   \label{fig:CARF}
\end{figure*}

\section{METHODOLOGY}

\label{sec:method}

The overall framework of CARF is illustrated in Fig.~\ref{fig:CARF}.
To realize the above objective, we first construct progress supervision
from successful trajectories in Section~\ref{sec:data}. We then train
a progress prediction scorer to estimate the contribution of action
segments in Section~\ref{sec:scorer}. Finally, the resulting importance
scores obtained from unseen failure trajectories guide the attraction-repulsion flow-matching objective in
Section~\ref{sec:flow}.
\subsection{Progress Score Generation for Success Trajectories} 
\label{sec:data}
To estimate the importance of each action step to final success, we assign a \textit{progress score} to each step for most tasks. We then derive the \textit{importance score} of a transition from the temporal difference between two consecutive progress scores, reflecting how much the action advances the task toward success based on current observation.
Considering a complex long-horizon task execution, the \textit{importance score} increases over time within each subtask, as later stages are closer to the subtask outcome, allow fewer feasible corrective actions, and tolerate smaller execution errors.



Building upon this, progress scores for each demonstration trajectory are automatically labeled. Specifically, we first divide long-horizon tasks into multiple subtasks based on \textit{change in the gripper state}, which typically indicates a meaningful transition in manipulation progress. 
Let a successful trajectory be segmented into $K$ subtasks.
We assign the $k$-th subtask a normalized progress interval
\begin{equation}
\mathcal{I}*k = [b*{k-1}, b_k],
\qquad
b_k = \frac{k}{K},
\quad k=1,\ldots,K.
\label{eq:progress_interval}
\end{equation}
Within each subtask, all steps preceding the last $t_0$ steps are assigned the lower bound $b_{k-1}$ of the corresponding progress interval. For the last $t_0$ steps, let $r_t\in[0,1]$ denote the normalized temporal position within this transition region. Their progress labels are defined as
\begin{equation}
p_t =
b_{k-1}
+
(b_k-b_{k-1})\phi_{\mathrm{exp}}(r_t),
\label{eq:progress_label}
\end{equation}
where
\begin{equation}
\phi_{\mathrm{exp}}(r_t)
=
\frac{\exp(\beta r_t)-1}{\exp(\beta)-1},
\qquad \beta>0,
\label{eq:exp_mapping}
\end{equation}
and $\phi_{\mathrm{exp}}:[0,1]\rightarrow[0,1]$ is a monotonically increasing exponential mapping satisfying
$\phi_{\mathrm{exp}}(0)=0$ and $\phi_{\mathrm{exp}}(1)=1$.
Thus, the progress label remains constant at $b_{k-1}$ for the early portion of each subtask and increases smoothly from $b_{k-1}$ to $b_k$ during its last $t_0$ steps.
This formulation guarantees monotonically increasing progress
within each subtask, while the interval width $1/K$ maintains a
consistent global progress range across tasks.
As a result, it guarantees importance scores to be also monotonous over time due to exponential labeling. 
The middle plot (`Labeled scores') in Figure~\ref{fig:CARF} (a) illustrates progress scores for a complete trajectory at each action step, where the red dashed lines indicate the boundaries between subtasks.

To expand the data distribution and prevent overfitting, we generate augmented trajectories by randomly selecting several action steps and perturbing gripper states at those time steps in successful demonstrations (referred to as `random state' in Figure~\ref{fig:CARF} (a)). Such perturbations typically cause the trajectory to fail at that point, and the progress score drops back to zero afterwards. Since the perturbation time steps are sampled randomly, a single successful trajectory can produce multiple failure variants, each with a different score profile. As a result, the training data of our scorer is mixed with action executions of real successes and synthetic failures, improving scorer's generalizability. 


\subsection{Importance Score Generation}
\label{sec:scorer}
After preparing the dataset, as illustrated in Figure~\ref{fig:CARF}(b), we train a progress prediction scorer to estimate the task progress associated with each action step in failed execution segments conditioned on the current observation. Specifically, given an observation and the corresponding H-step action at each timestep, the scorer predicts H step-wise progress scores (one for each action), and is supervised using the progress annotations introduced in Section~\ref{sec:data}.

Since flow matching policy operates over a multi-step action chunk, importance score of this action chunk is computed by the numerical difference of progress scores between the last and first timestep of this chunk. Specifically, we define the importance score of a chunk spanning
timesteps $t$ to $t+H-1$ as
\begin{equation}
    g(x_{t:t+H-1})
    =
    \hat{p}_{t+H-1}-\hat{p}_{t}.
    \label{eq:importance_score}
\end{equation}
Equivalently,
\begin{equation}
    g(x_{t:t+H-1})
    =
    \sum_{i=t}^{t+H-2}
    \left(\hat{p}_{i+1}-\hat{p}_{i}\right),
    \label{eq:importance_telescoping}
\end{equation}
showing that the chunk-level score measures the accumulated progress
over the corresponding action sequence rather than its absolute
progress value. This design is grounded on the monotonic increase of progress scores in the data labeling process (Section~\ref{sec:data}) in each successful task executions, so that the failure-critical action chunks can be clearly indicated with non-positive delta values.

The progress prediction scorer is first trained independently on the labeled successful trajectories and their augmented variants. After training, its parameters are frozen and the scorer is only used to provide importance scores for policy training. The scorer is not required during policy inference.

\subsection{Attraction-Repulsion Contrastive Flow Matching Training}
\label{sec:flow}
 We train flow matching policy on mixture of successful and real failed demonstrations instead of the augmented ones. During training, each failure trajectory is evaluated by the frozen scorer (Section~\ref{sec:scorer}) that produces an importance score for each action chunk, as exemplified in the lower-left plot of Figure~\ref{fig:CARF}(c). High values indicate useful action segments in failed task executions that the policy should learn (attraction), while low scores reflect undesirable behaviors that should be suppressed (repulsion). 
 
We incorporate these scores into policy training through adaptive losses to guide the training of our flow matching policy with both successful and failed demonstrations in a balanced way as shown in Figure~\ref{fig:CARF}(c). 
Specifically, we first form a mixed dataset with both success and failure data, where the failure data used here are not the artificially generated failures from Section~\ref{sec:data}, but come directly from the existing dataset. During training, the loss contributions of each failure trajectory segmentation are adaptively leveraged by the sequence importance score from the frozen progress prediction scorer. In this way, the policy is encouraged to focus on informative failure segments (\textit{Progressive Segments}) while suppressing misleading ones (\textit{Failure-Critical Segments}). 


Based on the standard interpolation process of flow matching with target velocity 
\begin{equation}
    v^*(x_t,t) = \frac{dx_t}{dt} = x_1 - x_0,
\end{equation}
we modify the flow matching objective conditioned on the scorer prediction, denoted as $g(x)$. With a predefined threshold, we first partition the failure data $x_{\text{fail}}$ into following categories:
\begin{equation}
\begin{aligned}
x_{\mathrm{suc}\mid\mathrm{fail}}
&= \left\{x \in x_{\text{fail}} \mid g(x) \ge \tau_1 \right\}, \\
x_{\mathrm{fail}\mid\mathrm{fail}}
&= \left\{x \in x_{\text{fail}} \mid g(x) \le \tau_2 \right\}. \\
x_{\mathrm{ind}}
&= \left\{x \in x_{\text{fail}} \mid \tau_2 < g(x) < \tau_1 \right\}. 
\end{aligned}
\end{equation}

Among them, two subsets correspond to distinct and informative categories within the failure dataset: Progressive Segments $x_{\mathrm{suc}\mid\mathrm{fail}}$, which contain high-quality behavioral segments despite originating from failed trajectories, and Failure-Critical Segments $x_{\mathrm{fail}\mid\mathrm{fail}}$, which represent genuinely incorrect behaviors that policy should avoid. Both of them provide supervision signals in the flow matching training, while Indeterminate Segments $x_{\mathrm{ind}}$ are discarded. Based on this partition, the attraction-repulsion objective at each training iteration is formulated as:
\begin{gather}
L_{\text{suc}}(\theta)
= \mathbb{E}_{(x,t)\sim x_{\text{suc}}}
\left[\left\|v^*(x_t,t)-v_{\theta}(x_t,t)\right\|^2\right], \\
L_{\text{suc}\mid\text{fail}}(\theta)
= \mathbb{E}_{(x,t)\sim x_{\text{suc}\mid\text{fail}}}
\left[\left\|v^*(x_t,t)-v_{\theta}(x_t,t)\right\|^2\right], \\
L_{\text{fail}\mid\text{fail}}(\theta)
= \mathbb{E}_{(x,t)\sim x_{\text{fail}\mid\text{fail}}}
\left[\left\|v^*(x_t,t)-v_{\theta}(x_t,t)\right\|^2\right], \\
\begin{aligned}
L(\theta)
={}& L_{\text{suc}}(\theta)
+ \lambda_1 L_{\text{suc}\mid\text{fail}}(\theta) \\
&+ \lambda_2\,\operatorname{ReLU}\!\Big(
\operatorname{stopgrad}\!\left(
L_{\text{suc}\mid\text{fail}}(\theta)
\right)
- L_{\text{fail}\mid\text{fail}}(\theta)
\Big).
\end{aligned}
\end{gather}
where $\lambda_1$ and $\lambda_2$ are the hyperparameters to control the strengths of attraction and repulsion respectively. 

 It is worth noting that directly
introducing a negative loss term, e.g.,
$-\mathcal{L}_{\mathrm{fail}\mid\mathrm{fail}}$, may lead to unstable
optimization, since the objective can be reduced by indefinitely
increasing the prediction error on failure-critical samples.
To avoid such unbounded repulsion, we instead use the positive
$\mathcal{L}_{\mathrm{suc}\mid\mathrm{fail}}$ as a dynamic reference
and constrain the repulsion through a ReLU margin.
Specifically, let
\begin{equation}
    c =
    \operatorname{stopgrad}
    \left(
        \mathcal{L}_{\mathrm{suc}\mid\mathrm{fail}}
    \right),
\end{equation}
and denote the repulsion term as
\begin{equation}
    \mathcal{R}(\theta)
    =
    \operatorname{ReLU}
    \left(
        c-\mathcal{L}_{\mathrm{fail}\mid\mathrm{fail}}(\theta)
    \right).
    \label{eq:repulsion_term}
\end{equation}
Since both the flow-matching losses and $\mathcal{R}(\theta)$ are
non-negative, the overall training objective remains lower bounded.
Moreover, except at the ReLU boundary, gradient in policy training process is
\begin{equation}
\nabla_{\theta}\mathcal{R}(\theta)
=
\begin{cases}
-\nabla_{\theta}
\mathcal{L}_{\mathrm{fail}\mid\mathrm{fail}}(\theta),
&
\mathcal{L}_{\mathrm{fail}\mid\mathrm{fail}} < c,
\\[3pt]
0,
&
\mathcal{L}_{\mathrm{fail}\mid\mathrm{fail}} \geq c.
\end{cases}
\label{eq:repulsion_gradient}
\end{equation}
Therefore, when the failure loss is smaller than the positive
reference, minimizing $\mathcal{R}$ explicitly increases the
flow-matching error on Failure-Critical Segments, producing the
desired repulsive gradient. Once the failure loss exceeds the
reference, the repulsive gradient vanishes. Meanwhile,
$\mathcal{L}_{\mathrm{suc}\mid\mathrm{fail}}$ typically decreases
as training progresses, yielding an increasingly conservative
repulsion margin. In this way, CARF pushes the policy away from
failure-critical behaviors only when necessary, while avoiding
unbounded separation and maintaining stable optimization.

\section{Experiments}
\label{sec:ex}
\subsection{Experimental Setup}
\begin{figure*}[t]
  \centering
   \includegraphics[width=0.92\textwidth]{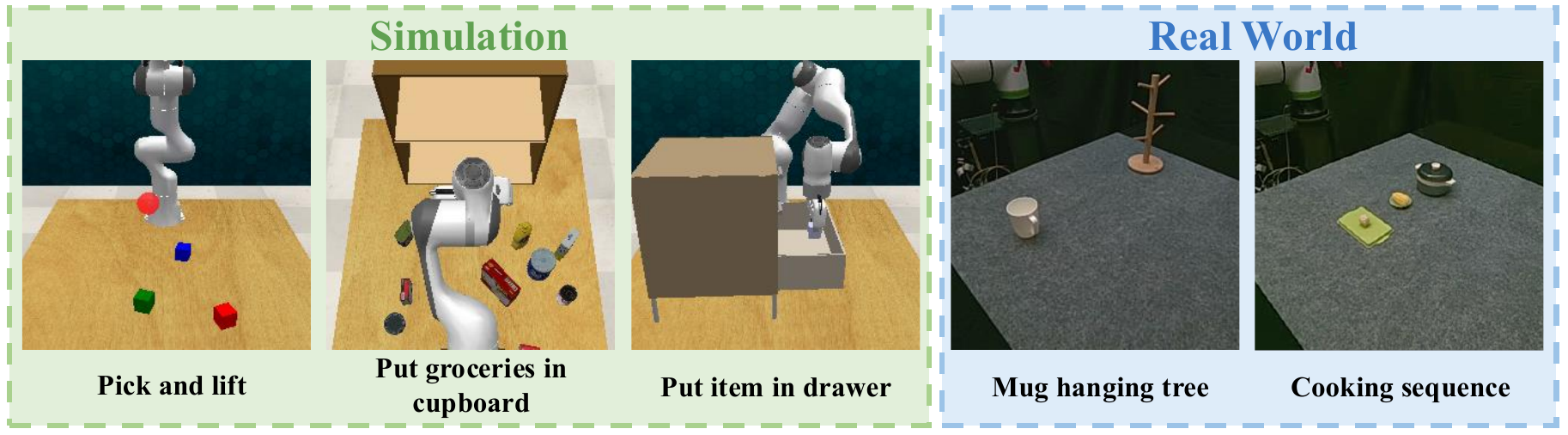}
   \caption{Environment visualization of tasks in simulation and real world.
   }
   \label{fig:task}
\end{figure*}

\textbf{Manipulation Tasks.}
As shown in Figure~\ref{fig:task}, we conduct experiments on three different tasks from RLBench~\cite{james2020rlbench} simulation and two tasks in the real world. In simulation, failure data are collected by perturbing successful trajectories at annotated keypoints, following \cite{duan2024aha}. In real world, failure data come from natural failures of human operators during teleoperation, which means we record both success and failure data in the data collection process.

Patterns of failure data in simulation and real world are intentionally different. Generated failed trajectories in simulation are allowed to continue after the failure, producing long failure sequences that may contain both misleading post-failure actions and partially useful motions. In contrast, real-world teleoperation failures terminate once the operator identifies an unrecoverable mistake. These two settings represent complementary types of imperfect robotic data: deployment failures with post-failure continuation, and human-collected failures with early termination. This is designed to show that CARF can improve performance in both settings, suggesting that the proposed attraction-repulsion mechanism is not tied to a specific failure collection protocol.

Besides, these tasks cover diverse manipulation modes for evaluating our method. \textit{Pick and Lift} requires moving the target red block to a marked position (red sphere). \textit{Put Groceries in Cupboard} requires placing a cereal box from a cluttered tabletop into a cupboard shelf. \textit{Put Item in Drawer} involves opening a drawer and placing an object inside. \textit{Mug Hanging Tree} requires hanging a mug by its handle on a pole. \textit{Cooking Sequence} evaluates long-horizon execution through an ordered sequence of opening the pot, placing corn inside, and putting broccoli on the plate. As shown in Table~\ref{table:task}, these tasks span different levels of dexterity and task complexity.
\begin{table}[t]
\caption{Capability requirements of designed robotics tasks.
}
\renewcommand{\arraystretch}{1.25}
\vspace{-2pt}
\centering
\footnotesize 
\setlength{\tabcolsep}{14pt}
\begin{threeparttable}
\resizebox{\columnwidth}{!}{
\begin{tabular}{c|c|c|c|c}
\toprule[1pt]
Tasks & Basic Interaction & Visual Object Detection & Precise Manipulation & Long Horizon \\
\midrule[0.6pt]

\textit{pick and Lift}
& \ding{51} & \ding{51} & \ding{55} & \ding{55}\\
\textit{put groceries in cupboard} & \ding{51} & \ding{51} & \ding{51} & \ding{55} \\

\textit{put item in drawer} & \ding{51} & \ding{55} & \ding{51} & \ding{51} \\
\textit{mug hanging tree}  & \ding{51} & \ding{55} & \ding{51} & \ding{55} \\
\textit{cooking sequence} & \ding{51} & \ding{55} & \ding{55} & \ding{51} \\
\bottomrule[1pt]
\end{tabular}
}
\end{threeparttable}
\label{table:task}
\end{table}

\textbf{Baselines.}
We compared our method with three different baselines:
\begin{itemize}
    \item \textbf{Success-only IL}~\cite{lipman2022flow}: We perform imitation learning using flow matching policy. With exactly the same network as our method, we only use success data to train and guide policy using normal MSE loss.
    \item \textbf{All-data IL}~\cite{lipman2022flow}: With everything else same as Success-only IL, we instead use both success and failure data to train imitation learning policy.
    \item \textbf{CGFM}~\cite{dhariwal2021diffusion}: We enhance flow matching policy with classifier guidance. Specifically, a binary (success/failure) classifier is trained based on trajectory-level labels in the mixed dataset. Then, it is used to guide flow matching policy during inference to lead policy to go to targeted area.
\end{itemize}
Each policy is evaluated for 100 independent rollouts per task for simulation, and 20 independent rollouts per task for real world. All methods use the same initial-condition distribution and success criterion. 

\textbf{Implementation Details.} For dataset preparation, we set $\beta=1$ and $t_0=30$ for all tasks. The progress prediction scorer is implemented as a three-layer MLP. Given the one-step observation feature and the corresponding sixteen-step action chunk, the scorer predicts sixteen scalar progress scores (one for each action) and is optimized using mean squared error (MSE) against the automatically generated progress labels.
We train the scorer independently for 10 epochs using AdamW with a learning rate of 3e-4 and a batch size of 64. Once trained, its parameters are frozen and the scorer is used only to assign importance scores during policy training and not required during policy inference.
For CARF policy learning, all methods use the same flow-matching architecture and training configuration for fair comparison. Policies are also optimized using AdamW with a learning rate of 1e-5, a batch size of 64, and 50 training epoches. 

For CARF parameters, we set
$\lambda_1=0.05$ and
$\lambda_2=0.01$ for the attraction and repulsion
objectives. These two hyperparameters are shared across all tasks and are
not tuned separately for individual environments or manipulation tasks.
The thresholds $\tau_1$ and $\tau_2$ are determined according to the scale of
the progress scores induced by the number of subtasks.
Since the overall progress-score range is divided across subtasks, tasks with
more subtasks have smaller per-subtask score increments and consequently use
smaller threshold values.
Therefore, $\tau_1$ and $\tau_2$ are adjusted according to the number of
subtasks rather than independently tuned based on policy performance.

\subsection{Experimental Results}
\begin{table*}[!ht]
\caption{Quantitative result comparison between \ours and other baselines.
}
\renewcommand{\arraystretch}{1.25}
\vspace{-2pt}
\centering
\footnotesize 
\setlength{\tabcolsep}{14pt}
\begin{threeparttable}
\resizebox{\textwidth}{!}{
\begin{tabular}{cccccc}
\toprule[1pt]
\multirow{3}{*}{Dataset \& Methods} & \multicolumn{3}{c}{Simulation} & \multicolumn{2}{c}{Real World}\\
\cmidrule(lr){2-4}\cmidrule(lr){5-6}
& \makecell{\textit{pick and lift}} & \makecell{\textit{put groceries}\\ \textit{in cupboard}} & \makecell{\textit{put item}\\\textit{in drawer}} & \makecell{\textit{mug hanging}\\\textit{tree}} & \makecell{\textit{cooking sequence}} \\
\midrule[0.6pt]
Success data & 200 & 200 & 200 & 100 & 50 \\
Failure data & 56 & 110 & 68 & 26 & 29 \\
\midrule[0.6pt]
Success-only IL~\cite{lipman2022flow}
& 0.38 & 0.23 & 0.62 & 0.50 & 0.55 \\
All-data IL~\cite{lipman2022flow}
& 0.32 (-15.8\%) & 0.20 (-13.0\%) & 0.53 (-14.5\%) & 0.35 (-30.0\%) & 0.45 (-18.2\%) \\

CGFM~\cite{dhariwal2021diffusion} & 0.24 (-36.8\%) & 0.14 (-39.1\%) & 0.47 (-24.2\%) & 0.45 (-10.0\%) & 0.40 (-27.3\%) \\
Ours & \textbf{0.48 (+26.3\%)} & \textbf{0.28 (+21.7\%)} & \textbf{0.69 (+11.3\%)} & \textbf{0.85 (+70.0\%)} & \textbf{0.70 (+27.3\%)} \\

\bottomrule[1pt]
\end{tabular}
}
\end{threeparttable}
\label{table:main}
\end{table*}


\begin{table}{r}
\centering
\vspace{-10pt}
\caption{Ablation study on attraction-repulsion items.}
\label{tab:ablation}
\resizebox{\linewidth}{!}{
\begin{tabular}{c|c|c|c}
\toprule
Task & \textit{pick and lift} & \textit{put grocery in cupboard} & \textit{mug hanging tree}\\
\midrule
Success-only IL & 0.38 & 0.23 & 0.50\\
IL \textit{w/} only positive terms & 0.39 & 0.21 & 0.70 \\
IL \textit{w/} only negative terms & 0.43 & 0.26 & 0.60\\
\midrule
CARF (ours) & \textbf{0.48} & \textbf{0.28} & \textbf{0.85}\\
\bottomrule
\end{tabular}}
\vspace{-0.2cm}
\end{table}
Table~\ref{table:main} shows the main results of our methods compared to the aforementioned baselines. 
Simply incorporating all failed trajectories into
imitation learning consistently degrades performance compared with
training on successful demonstrations only. This result indicates
that increasing the amount of training data does not necessarily
benefit policy learning when the additional demonstrations contain
heterogeneous and conflicting behaviors. In particular, failed
trajectories should not be uniformly treated as additional expert
demonstrations, motivating a more selective use of their informative
segments. This also supports our treatment of Indeterminate Segments: behaviors with ambiguous contribution to task progress may introduce unreliable supervision if they are directly included in policy optimization.
It is interesting to find out that CGFM actually reaches worse result than pure imitation learning. This reflects that classifier guidance is not suitable for this task which requires specific and precise distinguishment between different categories, especially with small dataset. In this case, the boundary between different categories is hard to recognize, which leads to wrong gradient during policy inference.

On the other hand, instead of directly fitting the limited training distribution or relying on unstable classifier gradients, our method explicitly exploits the usefulness of failed trajectories and selectively incorporates informative samples into policy learning. Therefore, it can improve data efficiency while reducing the risk of introducing misleading supervision from low-quality trajectories. More importantly, our results show that meaningful improvement can still be achieved by leveraging only a limited amount of imperfect data, instead of relying on large-scale demonstrations to learn an explicit boundary between correct and incorrect behaviors. 

Table~\ref{tab:ablation} presents ablation results to see the actual effect on attraction and repulsion design. From the table, we can see that for most of the time, attraction and repulsion will solely take relatively smaller effect on the result, but combine them together will improve the final results most for both failure setups. It demonstrate that attraction and repulsion play complementary roles: attraction extracts useful behavior from imperfect trajectories, while repulsion prevents the policy from imitating failure-critical actions. Their combination therefore enables more effective utilization of failed demonstrations than either component alone.


\begin{figure*}[!t]
    \centering

    \includegraphics[width=0.4\textwidth]{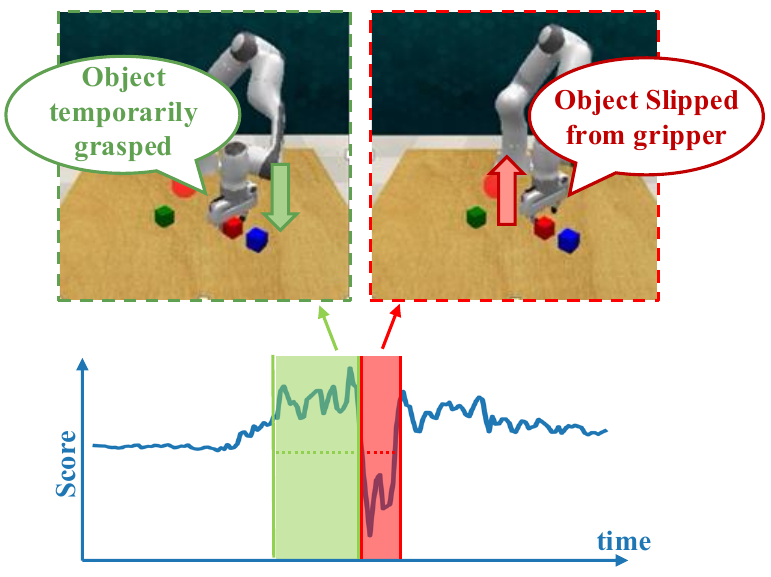}
    \includegraphics[width=0.42\textwidth]{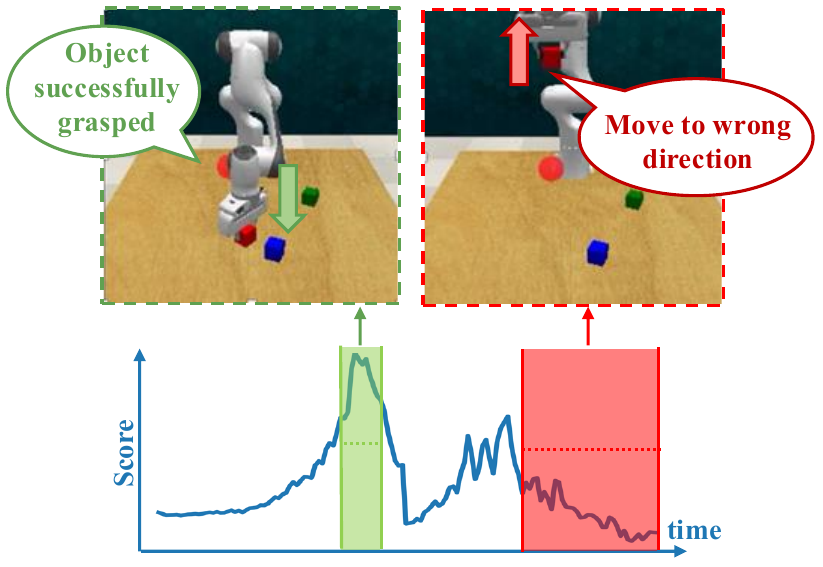}
    \includegraphics[width=0.4\textwidth]{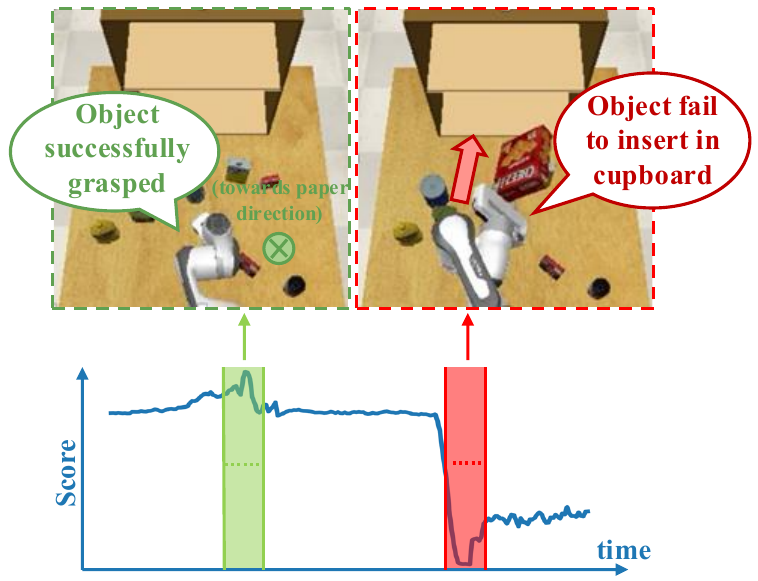}
    \includegraphics[width=0.42\textwidth]{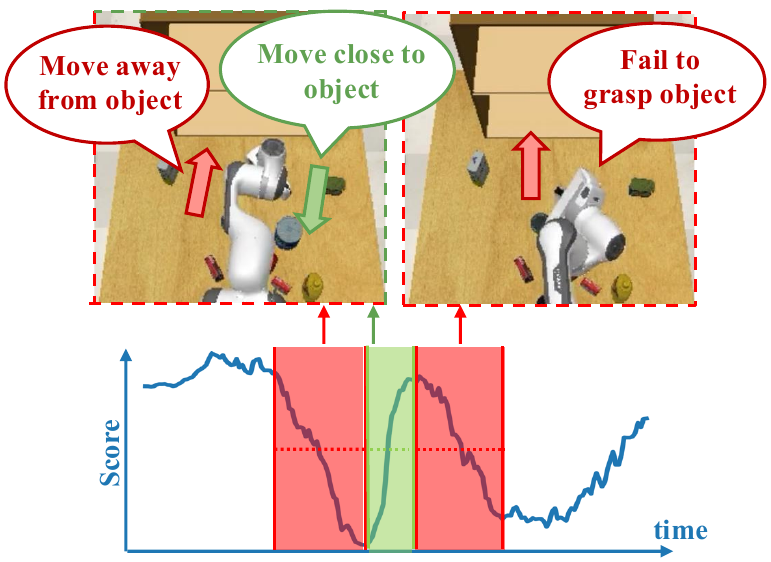}
    \caption{Scorer performance on representative failure trajectories. Green and Red respectively mean \textit{Progressive Segments} and \textit{Failure-Critical Segments}, and arrows in pictures mean current action moving direction of gripper end effector.}
    \label{fig:scorer}
\end{figure*}
\subsection{Scorer performance on failure trajectories}
\begin{figure*}[t]
  \centering
   \includegraphics[width=0.9\textwidth]{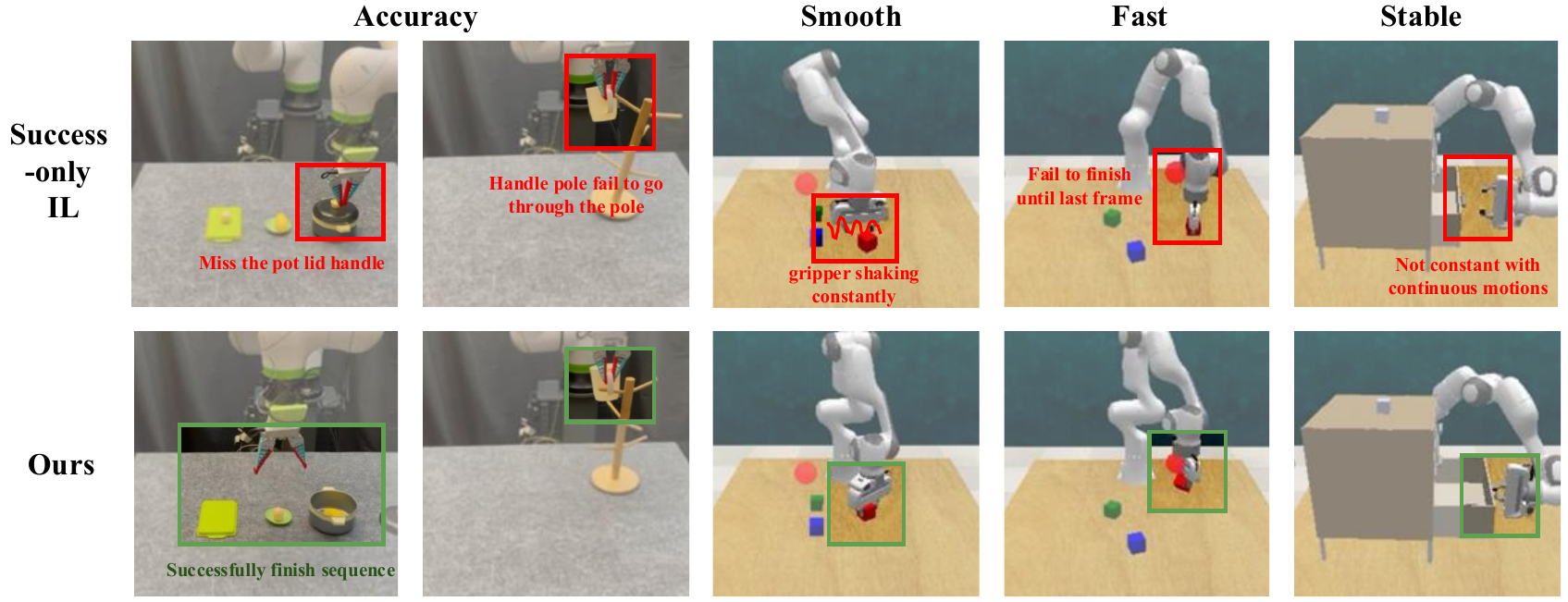}
   \caption{Visualization of failure improvement, where the improvement parts are annotated in the pictures using highlight rectangles with text descriptions.
   }
   \label{fig:difference}
   \vspace{-5pt}
\end{figure*}
To investigate whether our scorer can identify different trajectory types, we visualize the sequence-level importance scores derived from the scorer predictions within a failed trajectory, as shown in Figure~\ref{fig:scorer}. We observe that the importance scores soar upon the completion of a subtask, while they drop below zero when obvious errors occur or the trajectory starts deviating. This phenomenon is fully consistent with our expectation.
Notably, in some segments, although the action is incorrect when considered as a whole trajectory, it may still contain partially correct information. This can lead to a neutralized importance score close to zero, similar to the value observed when the agent is not close enough but still approaching to the target object. This explains why we only use the high-value and low-value regions for training. Such cases are inevitable in simulation tasks: even if a previous subtask fails, the policy continues to execute subsequent subtasks. In contrast, this situation does not happen in real-world tasks, where execution ends immediately when a failure is detected. This may also explain why our real-world results are slightly better since these misleading actions, mixed with correct motion patterns but actually caused by earlier failures, do not appear in real-world executions. This finding is consistent with this interpretation: early-terminated failures contain fewer misleading post-failure actions, making the separation between useful and harmful segments cleaner for the scorer.

\subsection{Visualization about failure improvement}

In this section, we compare the performance differences between our method and the Success-only IL baseline, as shown in Figure~\ref{fig:difference}. The results show that our method addresses several key weaknesses of imitation learning. \textbf{1) Accuracy.} Success-only IL can be confused by the scene and fail to perform precise motions, while our method correctly identifies, lifts and places the object into right place. \textbf{2) Smoothness.} Success-only IL often trembles during inference, preventing precise grasp execution. In contrast, our method produces smoother trajectories and completes the task successfully. \textbf{3) Speed.} Due to repeated grasping attempts, Success-only IL may fail to finish the task within a reasonable number of steps, while our method directly reaches and fetches the object. \textbf{4) Stability.} In continuous-contact tasks, Success-only IL can produce unstable motions that cause the target object to slip, whereas our method maintains stable contact until reaching the final destination.

Overall, these observations show that our method improves both effectiveness and execution quality. Compared with Success-only IL, our method is better at resolving ambiguous observations, generating smooth and stable motions, and completing tasks within fewer interaction steps. This indicates that scorer-guided learning can provide more robust demonstrations for complex manipulation, leading to policies that are better aligned with real execution requirements.

\section{Conclusion and Limitation}
\label{sec:conclusion}
Overall, CARF provides a practical way to use imperfect robotic data by separating \textit{Progressive Segments} and \textit{Failure-Critical Segments} in failed trajectories. Rather than relying on value bootstrapping or selectively reusing only progressive segments from failed trajectories, CARF introduces a progress-based scorer and a bounded attraction--repulsion objective to shape flow-matching policy learning. Across simulated and real-world manipulation tasks, CARF improves over imitation learning baselines, suggesting that segment-level failure guidance can improve data efficiency and execution robustness.

CARF still has several limitations. First, the current scorer uses
gripper-state transitions for subtask segmentation, which works well
for grasp-centric tasks but may be less suitable for continuous-contact manipulation. Future work can explore more general
segmentation cues based on contact, force, or visual progress.
Second, CARF adopts a conservative strategy that only utilizes
high-confidence Progressive and Failure-Critical Segments, while
discarding Indeterminate Segments with ambiguous supervision.
Although this reduces the risk of introducing misleading training
signals, potentially useful information contained in these intermediate
segments remains underutilized. Future work could incorporate
uncertainty-aware or soft weighting mechanisms to make more complete
use of imperfect demonstrations.
Finally, the current threshold selection still depends on the
task-specific scale of the progress representation, which requires
lightweight calibration according to the subtask structure.
Although the attraction and repulsion weights are shared across all
tasks, developing an adaptive or task-agnostic thresholding strategy
could further reduce manual design.


\bibliographystyle{IEEEtran}
\bibliography{main}  

@article{huang2025using,
  title={Using Non-Expert Data to Robustify Imitation Learning via Offline Reinforcement Learning},
  author={Huang, Kevin and Scalise, Rosario and Winston, Cleah and Agrawal, Ayush and Zhang, Yunchu and Baijal, Rohan and Grotz, Markus and Boots, Byron and Burchfiel, Benjamin and Itkina, Masha and others},
  journal={arXiv preprint arXiv:2510.19495},
  year={2025}
}

@article{intelligence2025pi,
  title={$\pi^*_{0.6}$: a VLA That Learns From Experience},
  author={Intelligence, Physical and Amin, Ali and Aniceto, Raichelle and Balakrishna, Ashwin and Black, Kevin and Conley, Ken and Connors, Grace and Darpinian, James and Dhabalia, Karan and DiCarlo, Jared and others},
  journal={arXiv preprint arXiv:2511.14759},
  year={2025}
}

@inproceedings{xu2022discriminator,
  title={Discriminator-weighted offline imitation learning from suboptimal demonstrations},
  author={Xu, Haoran and Zhan, Xianyuan and Yin, Honglei and Qin, Huiling},
  booktitle={International Conference on Machine Learning},
  pages={24725--24742},
  year={2022},
  organization={PMLR}
}

@article{dass2025datamil,
  title={Datamil: Selecting data for robot imitation learning with datamodels},
  author={Dass, Shivin and Khaddaj, Alaa and Engstrom, Logan and Madry, Aleksander and Ilyas, Andrew and Mart{\'\i}n-Mart{\'\i}n, Roberto},
  journal={arXiv preprint arXiv:2505.09603},
  year={2025}
}

@article{du2023behavior,
  title={Behavior retrieval: Few-shot imitation learning by querying unlabeled datasets},
  author={Du, Maximilian and Nair, Suraj and Sadigh, Dorsa and Finn, Chelsea},
  journal={arXiv preprint arXiv:2304.08742},
  year={2023}
}

@article{lin2024flowretrieval,
  title={Flowretrieval: Flow-guided data retrieval for few-shot imitation learning},
  author={Lin, Li-Heng and Cui, Yuchen and Xie, Amber and Hua, Tianyu and Sadigh, Dorsa},
  journal={arXiv preprint arXiv:2408.16944},
  year={2024}
}

@article{zhao2023learning,
  title={Learning fine-grained bimanual manipulation with low-cost hardware},
  author={Zhao, Tony Z and Kumar, Vikash and Levine, Sergey and Finn, Chelsea},
  journal={arXiv preprint arXiv:2304.13705},
  year={2023}
}

@inproceedings{wu2025learning,
  title={Learning from imperfect demonstrations with self-supervision for robotic manipulation},
  author={Wu, Kun and Liu, Ning and Zhao, Zhen and Qiu, Di and Li, Jinming and Che, Zhengping and Xu, Zhiyuan and Tang, Jian},
  booktitle={2025 IEEE International Conference on Robotics and Automation (ICRA)},
  pages={16899--16906},
  year={2025},
  organization={IEEE}
}

@inproceedings{stoica2025contrastive,
  title={Contrastive flow matching},
  author={Stoica, George and Ramanujan, Vivek and Fan, Xiang and Farhadi, Ali and Krishna, Ranjay and Hoffman, Judy},
  booktitle={Proceedings of the IEEE/CVF International Conference on Computer Vision},
  pages={1185--1194},
  year={2025}
}

@article{lipman2022flow,
  title={Flow matching for generative modeling},
  author={Lipman, Yaron and Chen, Ricky TQ and Ben-Hamu, Heli and Nickel, Maximilian and Le, Matt},
  journal={arXiv preprint arXiv:2210.02747},
  year={2022}
}

@article{chi2025diffusion,
  title={Diffusion policy: Visuomotor policy learning via action diffusion},
  author={Chi, Cheng and Xu, Zhenjia and Feng, Siyuan and Cousineau, Eric and Du, Yilun and Burchfiel, Benjamin and Tedrake, Russ and Song, Shuran},
  journal={The International Journal of Robotics Research},
  volume={44},
  number={10-11},
  pages={1684--1704},
  year={2025},
  publisher={Sage Publications Sage UK: London, England}
}

@article{zhao2025dexh2r,
  title={Dexh2r: Task-oriented dexterous manipulation from human to robots},
  author={Zhao, Shuqi and Zhu, Xinghao and Chen, Yuxin and Li, Chenran and Xie, Yichen and Zhang, Xiang and Ding, Mingyu and Tomizuka, Masayoshi},
  journal={IEEE/ASME Transactions on Mechatronics},
  year={2025},
  publisher={IEEE}
}

@article{xu2025seeing,
  title={Seeing to Act, Prompting to Specify: A Bayesian Factorization of Vision Language Action Policy},
  author={Xu, Kechun and Zhu, Zhenjie and Chen, Anzhe and Zhao, Shuqi and Huang, Qing and Yang, Yifei and Lu, Haojian and Xiong, Rong and Tomizuka, Masayoshi and Wang, Yue},
  journal={arXiv preprint arXiv:2512.11218},
  year={2025}
}

@article{zhao2025dexctrl,
  title={Dexctrl: Towards sim-to-real dexterity with adaptive controller learning},
  author={Zhao, Shuqi and Yang, Ke and Chen, Yuxin and Li, Chenran and Xie, Yichen and Zhang, Xiang and Wang, Changhao and Tomizuka, Masayoshi},
  journal={arXiv preprint arXiv:2505.00991},
  year={2025}
}

@article{kostrikov2021offline,
  title={Offline reinforcement learning with implicit q-learning},
  author={Kostrikov, Ilya and Nair, Ashvin and Levine, Sergey},
  journal={arXiv preprint arXiv:2110.06169},
  year={2021}
}

@article{duan2024aha,
  title={Aha: A vision-language-model for detecting and reasoning over failures in robotic manipulation},
  author={Duan, Jiafei and Pumacay, Wilbert and Kumar, Nishanth and Wang, Yi Ru and Tian, Shulin and Yuan, Wentao and Krishna, Ranjay and Fox, Dieter and Mandlekar, Ajay and Guo, Yijie},
  journal={arXiv preprint arXiv:2410.00371},
  year={2024}
}

@article{james2020rlbench,
  title={Rlbench: The robot learning benchmark \& learning environment},
  author={James, Stephen and Ma, Zicong and Arrojo, David Rovick and Davison, Andrew J},
  journal={IEEE Robotics and Automation Letters},
  volume={5},
  number={2},
  pages={3019--3026},
  year={2020},
  publisher={IEEE}
}

@article{yan2026progressvla,
  title={Progressvla: Progress-guided diffusion policy for vision-language robotic manipulation},
  author={Yan, Hongyu and Li, Qiwei and Yang, Jiaolong and Mu, Yadong},
  journal={arXiv preprint arXiv:2603.27670},
  year={2026}
}

@article{zheng2026rewind,
  title={Rewind-IL: Online failure detection and state respawning for imitation learning},
  author={Zheng, Gehan and Seenivasan, Sanjay and Johnson-Roberson, Matthew and Zhi, Weiming},
  journal={arXiv preprint arXiv:2604.16683},
  year={2026}
}

@article{huang2025fail2progress,
  title={Fail2progress: Learning from real-world robot failures with stein variational inference},
  author={Huang, Yixuan and Alvina, Novella and Shanthi, Mohanraj Devendran and Hermans, Tucker},
  journal={arXiv preprint arXiv:2509.01746},
  year={2025}
}

@inproceedings{mao2026beyond,
  title={Beyond Success: Refining Elegant Robot Manipulation from Mixed-Quality Data via Just-in-Time Intervention},
  author={Mao, Yanbo and Fu, Jianlong and Zhang, Ruoxuan and Xie, Hongxia and Yao, Meibao},
  booktitle={Proceedings of the IEEE/CVF Conference on Computer Vision and Pattern Recognition},
  pages={13508--13518},
  year={2026}
}

@inproceedings{kim2021demodice,
  title={Demodice: Offline imitation learning with supplementary imperfect demonstrations},
  author={Kim, Geon-Hyeong and Seo, Seokin and Lee, Jongmin and Jeon, Wonseok and Hwang, HyeongJoo and Yang, Hongseok and Kim, Kee-Eung},
  booktitle={International Conference on Learning Representations},
  year={2021}
}

@article{zheng2026failing,
  title={Failing Forward: Adaptive Failure-Informed Learning for Vision-Language-Action Models},
  author={Zheng, Meng and Marri, Samhita and Choudhuri, Anwesa and Planche, Benjamin and Gao, Zhongpai and Nguyen, Van Nguyen and Chen, Terrence and Chowdhary, Girish and Wu, Ziyan},
  journal={arXiv preprint arXiv:2605.08434},
  year={2026}
}

@article{giridhar2026beyond,
  title={Beyond Imitation: Self-Improving Robot Policies via Off-Policy Q-Planning},
  author={Giridhar, Varun and Khandelwal, Anant and Collins, Jeremy A and Georgiev, Ignat and Garg, Animesh},
  journal={arXiv preprint arXiv:2608.21204},
  year={2026}
}

@article{hao2026far,
  title={FAR: Failure-Aware Retry for Test-Time Recovery and Continual Policy Improvement},
  author={Hao, Haoran and Syed, Shahram Najam and Ichnowski, Jeffrey and Schneider, Jeff},
  journal={arXiv preprint arXiv:2607.01111},
  year={2026}
}

@article{wei2026ambient,
  title={Ambient Diffusion Policy: Imitation Learning from Suboptimal Data in Robotics},
  author={Wei, Adam and Pfaff, Nicholas and Cohn, Thomas and Day{\i}, Arif Kerem and Daskalakis, Constantinos and Daras, Giannis and Tedrake, Russ},
  journal={arXiv preprint arXiv:2606.12365},
  year={2026}
}

@article{kumar2020conservative,
  title={Conservative q-learning for offline reinforcement learning},
  author={Kumar, Aviral and Zhou, Aurick and Tucker, George and Levine, Sergey},
  journal={Advances in neural information processing systems},
  volume={33},
  pages={1179--1191},
  year={2020}
}

@article{hansen2023idql,
  title={Idql: Implicit q-learning as an actor-critic method with diffusion policies},
  author={Hansen-Estruch, Philippe and Kostrikov, Ilya and Janner, Michael and Kuba, Jakub Grudzien and Levine, Sergey},
  journal={arXiv preprint arXiv:2304.10573},
  year={2023}
}

@inproceedings{robomimic2021,
  title={What Matters in Learning from Offline Human Demonstrations for Robot Manipulation},
  author={Ajay Mandlekar and Danfei Xu and Josiah Wong and Soroush Nasiriany and Chen Wang and Rohun Kulkarni and Li Fei-Fei and Silvio Savarese and Yuke Zhu and Roberto Mart\'{i}n-Mart\'{i}n},
  booktitle={Conference on Robot Learning (CoRL)},
  year={2021}
}

@article{xie2026multi,
  title={Multi-Camera View Scaling for Data-Efficient Robot Imitation Learning},
  author={Xie, Yichen and Wang, Yixiao and Zhao, Shuqi and Wu, Cheng-En and Tomizuka, Masayoshi and Xie, Jianwen and Fang, Hao-Shu},
  journal={arXiv preprint arXiv:2604.00557},
  year={2026}
}

@article{chen2026craft,
  title={CRAFT: Video Diffusion for Bimanual Robot Data Generation},
  author={Chen, Jason and Liu, I and Sukhatme, Gaurav and Seita, Daniel and others},
  journal={arXiv preprint arXiv:2604.03552},
  year={2026}
}

@article{chi2024universal,
  title={Universal manipulation interface: In-the-wild robot teaching without in-the-wild robots},
  author={Chi, Cheng and Xu, Zhenjia and Pan, Chuer and Cousineau, Eric and Burchfiel, Benjamin and Feng, Siyuan and Tedrake, Russ and Song, Shuran},
  journal={arXiv preprint arXiv:2402.10329},
  year={2024}
}

@article{xu2025dexumi,
  title={Dexumi: Using human hand as the universal manipulation interface for dexterous manipulation},
  author={Xu, Mengda and Zhang, Han and Hou, Yifan and Xu, Zhenjia and Fan, Linxi and Veloso, Manuela and Song, Shuran},
  journal={arXiv preprint arXiv:2505.21864},
  year={2025}
}

@article{fang2025dexop,
  title={Dexop: A device for robotic transfer of dexterous human manipulation},
  author={Fang, Hao-Shu and Romero, Branden and Xie, Yichen and Hu, Arthur and Huang, Bo-Ruei and Alvarez, Juan and Kim, Matthew and Margolis, Gabriel and Anbarasu, Kavya and Tomizuka, Masayoshi and others},
  journal={arXiv preprint arXiv:2509.04441},
  year={2025}
}

@article{dhariwal2021diffusion,
  title={Diffusion models beat gans on image synthesis},
  author={Dhariwal, Prafulla and Nichol, Alexander},
  journal={Advances in neural information processing systems},
  volume={34},
  pages={8780--8794},
  year={2021}
}

@inproceedings{xia2025cage,
  title={Cage: Causal attention enables data-efficient generalizable robotic manipulation},
  author={Xia, Shangning and Fang, Hongjie and Lu, Cewu and Fang, Hao-Shu},
  booktitle={2025 IEEE International Conference on Robotics and Automation (ICRA)},
  pages={13242--13249},
  year={2025},
  organization={IEEE}
}

@article{zheng2026egoscale,
  title={Egoscale: Scaling dexterous manipulation with diverse egocentric human data},
  author={Zheng, Ruijie and Niu, Dantong and Xie, Yuqi and Wang, Jing and Xu, Mengda and Jiang, Yunfan and Casta{\~n}eda, Fernando and Hu, Fengyuan and Tan, You Liang and Fu, Letian and others},
  journal={arXiv preprint arXiv:2602.16710},
  year={2026}
}

@article{zheng2024data,
  title={Data Scaling Laws for Imitation Learning-Based End-to-End Autonomous Driving},
  author={Zheng, Yupeng and Yang, Pengxuan and Xia, Zhongpu and Zhang, Qichao and Zheng, Yuhang and Gu, Songen and Jin, Bu and Zhang, Teng and Lu, Ben and Han, Chao and others},
  journal={arXiv preprint arXiv:2412.02689},
  year={2024}
}

\end{document}